\documentclass[letterpaper, 10 pt, conference]{ieeeconf}  

\IEEEoverridecommandlockouts                              

\usepackage{graphics} 
\usepackage{epsfig} 
\usepackage{mathptmx} 
\usepackage{times} 
\usepackage{amsmath} 
\usepackage{amssymb}  

\usepackage{multirow} 
\usepackage{booktabs}
\usepackage{xcolor}

\newif\ifrevfinal

\revfinaltrue

\ifrevfinal
  \newcommand{\rev}[1]{#1}
\else
  \newcommand{\rev}[1]{\textcolor{red}{#1}}
\fi
\newcommand{\consistencyrev}[1]{{#1}}

\usepackage{caption}
\newcommand{\figurename}{Fig.}

\usepackage{cite}                
\usepackage[hidelinks]{hyperref}

\newif\ifblind

\title{\LARGE \bf
\rev{Improving Imitation Learning Efficiency for \\Manipulation through Geometric Prior Pretraining}
}

\ifblind
  \author{Anonymous Author(s)}
\else
    \IEEEoverridecommandlockouts    

    \author{Shogo Iwakata$^{1}$, Tomohiro Motoda$^{2,*}$, Ryosuke Yamada$^{3}$, Koshi Makihara$^{2}$, Ryoichi Nakajo$^{2}$, \\Keitaro Tanaka$^{5}$, Masaki Murooka$^{4}$,  
    Roman Mykhailyshyn$^{2}$, 
    Hirokatsu Kataoka$^{3,6}$, Shigeo Morishima$^{5}$, \\and Yukiyasu Domae$^{2}$
    \thanks{$^{1}$Waseda University}%
    \thanks{$^{2}$Embodied AI Research Team, AIRC and National Institute of Advanced Industrial Science and Technology (AIST)}
    \thanks{$^{3}$Computer Vision Research Team, AIRC and National Institute of Advanced Industrial Science and Technology (AIST)}
    \thanks{$^{4}$CNRS-AIST JRL (Joint Robotics Laboratory), IRL and National Institute of Advanced Industrial Science and Technology (AIST)}
    \thanks{$^{5}$Waseda Research Institute for Science and Engineering}
    \thanks{$^{6}$Visual Geometry Group, University of Oxford}
    \thanks{$^*$Corresponding Author}
    }
\fi

\begin{document}


\maketitle
\thispagestyle{empty}
\pagestyle{empty}

\begin{abstract}
\rev{Applying an imitation learning policy to a new manipulation task usually requires collecting new demonstrations and retraining the model, which makes sample efficiency a practical concern. Pretraining on large-scale robot datasets is effective in this respect, but such datasets are costly to collect and train on, while data augmentation techniques typically require a new round of data generation and retraining for each task. A complementary question is what useful prior can be provided to a policy at negligible cost before any task-specific data are collected. In this study, we construct a geometric visual pretraining dataset in which each scene contains only a plane, an object, and a hand, and trajectories are generated automatically. The scenes contain neither textures nor backgrounds; pretraining primarily exposes the policy to the geometric relationship between the hand and the object. Furthermore, representing the hand as a cube avoids tailoring the dataset to a specific robot morphology. We evaluate this geometric prior using ACT on three simulated robots across five manipulation tasks each, as well as on three real-world robot tasks. Across many of these robot--task combinations, fine-tuning from the geometric prior achieves higher success rates in the early stages of training than training from scratch while using only a small number of task demonstrations. These results suggest that even highly simplified geometric scenes can provide a useful initialization that transfers across robots and to real-world tasks when task data are limited.} 
\end{abstract}

\section{INTRODUCTION}

Robots capable of substituting for human labor are crucial 
 for reducing human burden and saving time~\cite{Intahchomphoo_Millar_Gundersen_Tschirhart_Meawasige_Salemi_2024}. 
Imitation learning has emerged as a promising approach
 for enabling robots to acquire complex skills, 
 as it bypasses costly reward engineering 
 and instead relies on human expert demonstrations~\cite{10.1016/j.robot.2008.10.024, Osa_2018}.
 \rev{Although recent imitation-learning policies have achieved impressive performance,
 their sample efficiency strongly depends on the task, the observation space, and the
 policy architecture~\cite{10.1145/3054912, Chen2022AnEI,finn2017oneshotvisualimitationlearning}.
 In practice, collecting task-specific demonstrations remains a major bottleneck when adapting
 policies to new tasks, and is often prohibitively expensive and time-consuming in real-world
 industrial and research settings.}
It is therefore critical to develop methods that can train accurate models efficiently using only a small amount of real data.

\begin{figure}[!t]
    \centering
    \includegraphics[width=0.95\columnwidth]{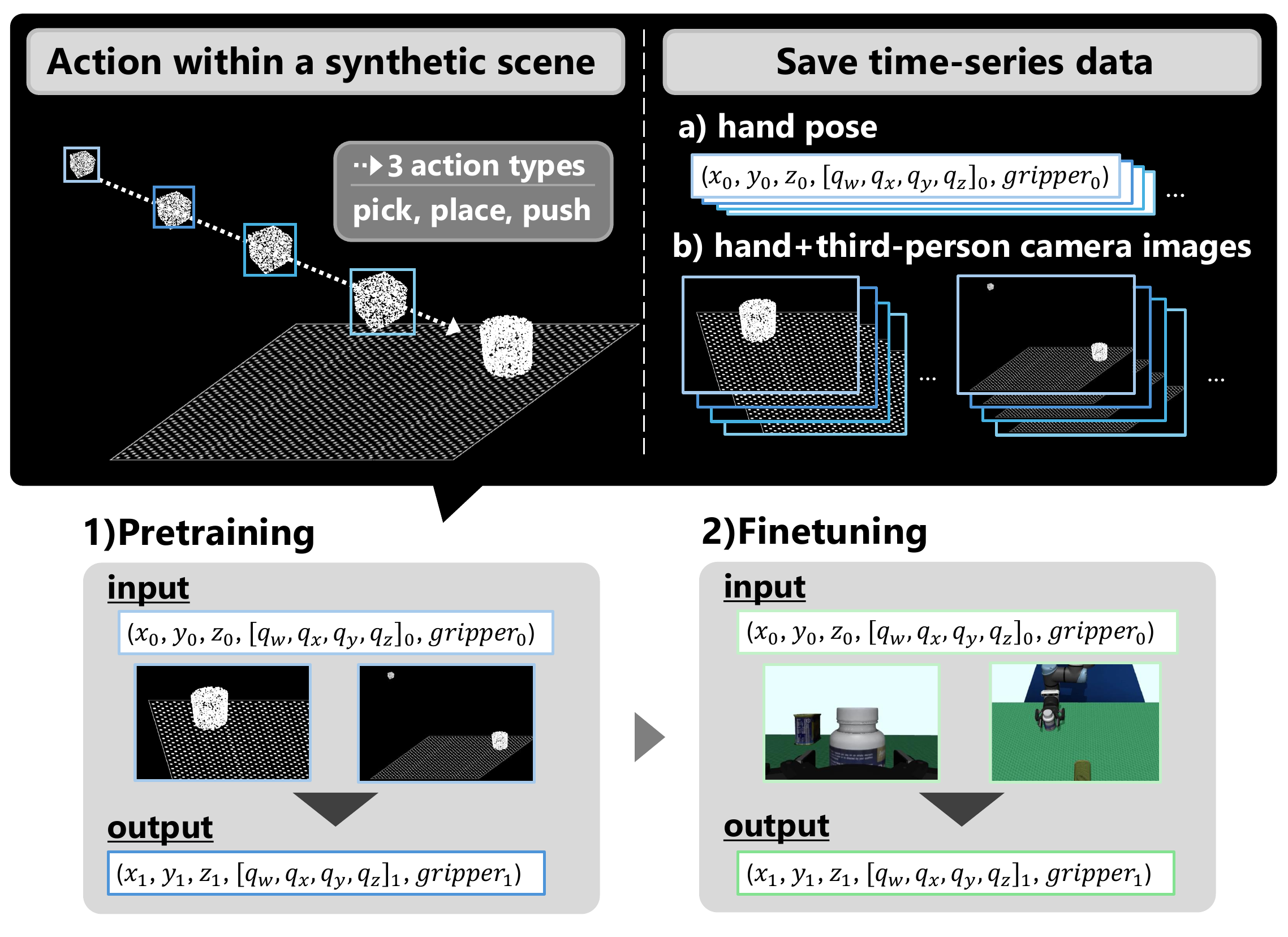}
    \caption{\textbf{An overview of the pretraining and imitation learning processes that utilize automatically generated synthetic data.}
    We first pretrain on synthetic data and then fine-tune on task-specific data. The synthetic data are generated in a simple scene with only a plane, an object, and a hand, \rev{in which motions that expose the relative geometry between the hand and the object are automatically reproduced as temporal sequences.}}
    \vspace{-2mm}
    \label{fig:abst}
\end{figure}

One promising direction to address this data constraint is to leverage large-scale, publicly available datasets collected from multiple robots and environments~\cite{dasari2020robonetlargescalemultirobotlearning, walke2023bridgedata, open_x_embodiment_rt_x_2023, khazatsky2024droid}.
\rev{Pretraining on large-scale data has been reported to let robots acquire broadly applicable skills,}
 and fine-tuning with a small amount of task-specific real data
 can improve accuracy and accelerate convergence~\cite{lin2024data, 9b18aa5285844beb9bb4620973805743, DBLP:journals/corr/abs-2405-14093}.
However, large-scale pretraining requires substantial computational resources, and its performance strongly depends on the selection and integration of data from different tasks, robots, and environments~\cite{10.5555/3666122.3669646}.
Another line of research focuses on data augmentation methods that generate additional demonstrations from limited task-specific data~\cite{DomainRandomization, mimicgen, IntervenGen, DemoGen, DreamGen,CP-gen}. 
These approaches can improve robustness and data efficiency within a given task.
However, they primarily enrich task-specific distributions \rev{rather than providing a task-agnostic starting point}, often requiring data generation and retraining for each new task or setting.
\rev{These issues motivate pretraining signals that are independent of any particular downstream task and that can be obtained without collecting additional robot data.}

To address these challenges, we propose an approach that generates \rev{purely geometric} synthetic data in a simplified scene and leverages it to pretrain the model\rev{, giving the policy what we call a geometric prior} (Fig.~\ref{fig:abst}).
In robotic manipulation, the relationship between the manipulated object and the manipulating hand is crucial for successful task execution~\cite{article_relativeposition}. 
Our key insight is to leverage this principle.
The proposed data are generated in an environment composed of only three elements: an object, a hand, and a plane (corresponding to the table in tabletop manipulation tasks).

\rev{Within this simplified setting, we automatically generate short time-series trajectories in which the hand approaches the object, aligns with it, and transports it to a target pose (Fig.~\ref{fig:abst}, top-right); the concrete motion types we instantiate are described in Sec.~\ref{sec:data}.} Through this design, when used for training, the proposed synthetic data restricts the model input to only these relative spatial relationships (e.g., the position and orientation between the object and the hand).
\rev{This design choice removes extraneous factors (e.g., varying backgrounds or lighting conditions) found in real-world data from the pretraining signal, so that the structure available to the model during pretraining is the relative geometry between the hand and the object. Because we do not assume any specific robot morphology, the proposed dataset abstracts hand position and orientation as a cube.}
Furthermore, \rev{object shape diversity is introduced} by varying primitive types and geometric parameters (e.g., size and proportions) used during generation.
Each data sample includes time-series information of hand poses, hand camera images, and third-person camera images. We evaluated the effectiveness of the proposed dataset through experiments conducted across simulation and real-world environments, involving multiple robots and tasks.
The results demonstrate that pretraining with the proposed data \rev{consistently improved early-stage learning across many robot--task combinations}. \rev{Our goal is not to replace large-scale robot datasets or task-specific data augmentation methods. Instead, we investigate whether simple geometric synthetic data can serve as an inexpensive pretraining signal that improves learning efficiency in modern visuomotor policies.}

\section{RELATED WORK}

\subsection{Large-scale Datasets and Generalist Robot Policies}

Recent robot-learning research has focused on collecting large-scale datasets and training generalist policies that operate across multiple tasks and embodiments.
Representative efforts include multi-robot datasets such as RoboNet~\cite{dasari2020robonetlargescalemultirobotlearning}, aggregated collections such as BridgeData V2~\cite{walke2023bridgedata}, and teleoperated datasets that emphasize environmental diversity, such as DROID~\cite{khazatsky2024droid}. 
Similarly, the RT series (RT-1~\cite{rt12022arxiv}, RT-2~\cite{zitkovich2023rt}, RT-X~\cite{open_x_embodiment_rt_x_2023}) demonstrates that large-scale robot and web-derived vision--language data can improve instruction following and high-level reasoning.
Studies including OpenVLA~\cite{kim24openvla}, Octo~\cite{octo_2023}, RoboAgent~\cite{bharadhwaj2023roboagentgeneralizationefficiencyrobot}, and What Matters in Learning from Large-Scale Datasets~\cite{saxena2025what} further analyze the effects of dataset composition, supervision, and training scale on generalization.

However, these approaches typically require extensive human demonstrations and substantial computational resources. Instead, we investigate whether \rev{geometric prior} pretraining can provide useful inductive biases at a lower cost. By constructing simple environments that emphasize relative geometric relationships critical for manipulation, our method aims to obtain pretraining benefits without relying on massive real-world datasets.

\subsection{Data Augmentation for Robotic Manipulation}

Data augmentation methods improve data efficiency and generalization by generating additional training samples from limited demonstrations.
Early approaches rely on low-level perturbations such as Domain Randomization~\cite{DomainRandomization}, while recent methods generate task-consistent action data.

MimicGen~\cite{mimicgen} recombines object-centric interactions from a small set of demonstrations to synthesize large-scale datasets. IntervenGen~\cite{IntervenGen} generates corrective intervention data to improve robustness, while DemoGen~\cite{DemoGen} creates new visuomotor demonstrations by modifying object configurations in 3D space. DreamGen~\cite{DreamGen} leverages image- and video-generation models to synthesize robot trajectories, and CP-Gen~\cite{CP-gen} generates demonstrations that satisfy task constraints across varying object configurations via optimization and motion planning.

Although effective, these methods generally depend on task-specific demonstrations, interventions, or manually defined task structures, making the generated distributions closely tied to particular tasks and environments. In contrast, we propose a task-agnostic pretraining framework \rev{that does not require task-specific demonstrations}, aiming to improve learning efficiency across diverse downstream manipulation tasks through a single pretraining stage.

\begin{figure*}[t]
    \centering
    \includegraphics[width=0.95\textwidth]{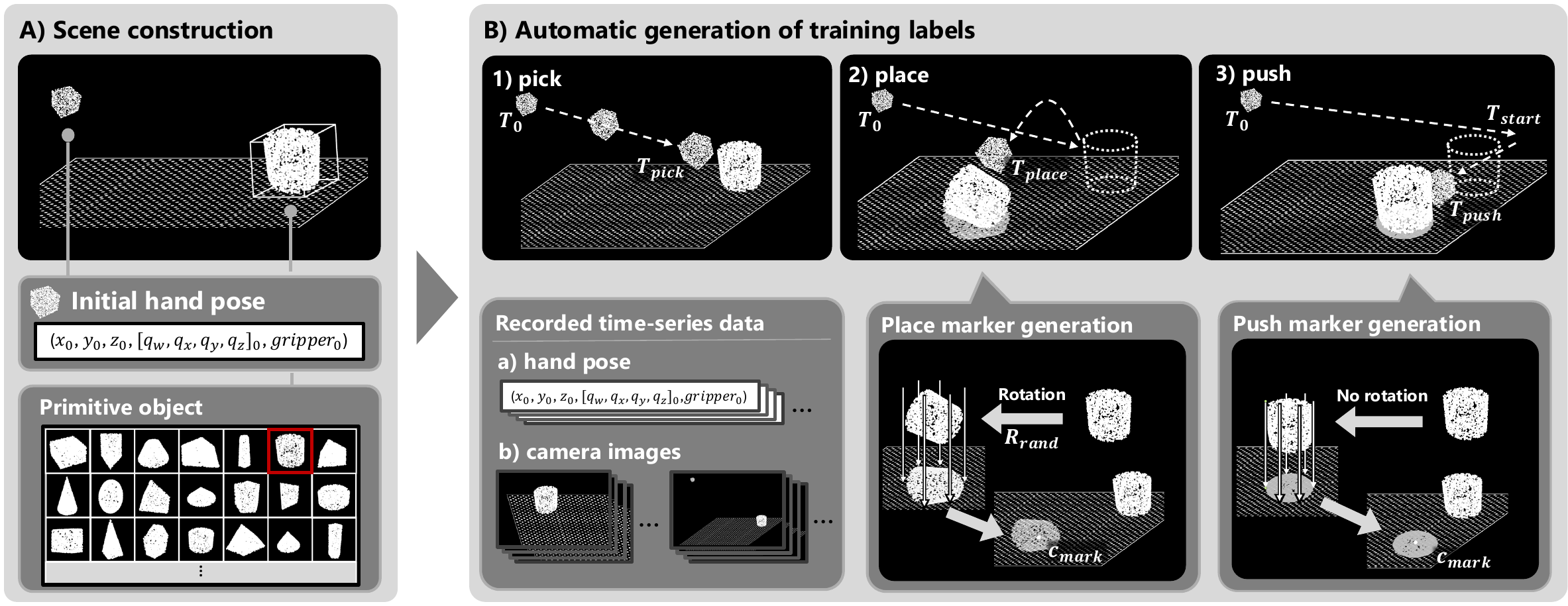}
    \caption{\textbf{Overview of Data Creation}. We generate a simple scene consisting of a plane, an object, and a hand. For the place and push tasks, the target pose is visually represented by a pattern placed on a plane, which we refer to as a marker. The marker is created by projecting the shapes of the objects onto the plane. Action trajectories for pick, place, and push are reproduced within this scene, and the information recorded during these actions is stored as time-series data.}
    \label{fig:data-creation}
    \vspace{-2mm}
\end{figure*}

\section{\rev{CREATION OF GEOMETRIC PRIOR DATA FOR PRETRAINING}}
\label{sec:data}

We automatically generate synthetic data in a purely geometric synthetic scene without physics for pretraining robot manipulation tasks. The synthetic scene consists of a floor plane, an object, and a hand, and reproduces three types of hand actions: pick, place, and push. The overview diagram of the data creation flow is shown in \figurename~\ref{fig:data-creation}. In this section, we describe in detail the construction of the synthetic scene and the method for automatically generating pretraining data (observations and hand trajectories) for each motion.

\subsection{Scene Construction} 
The simulation scenes are constructed entirely from point clouds. 
Each scene consists of a floor plane, one object, and a hand (\figurename~\ref{fig:data-creation}, top-left). 
To introduce variation, the floor size, floor height, and grid resolution are sampled with slight randomness. 
Objects are modeled as simple geometric primitives, such as prisms, pyramids, cylinders, cones, spheres, and truncated variants (\figurename~\ref{fig:data-creation} bottom left). 
To increase diversity, object sizes are randomly sampled over a wide range, regular polygon bases are slightly distorted, and some objects are slightly stretched along different axes (e.g., spheres become ellipsoids).
Each object is randomly positioned on the floor.
In addition, the hand is randomly placed at a height above the plane, and its orientation is set to point toward a position near the object so that the object appears in the hand camera at the initial pose.

\subsection{Automatic Generation of Training Labels}
For each demonstration, we generate a time-ordered sequence of data.
At every time step $t$, we record a tuple consisting of (i) the hand pose, (ii) the hand-camera image, and (iii) a third-person camera image.
The hand pose is represented by its position and orientation, and the gripper value is recorded as 0 or 1 to indicate open and closed states.

The length of each demonstration is determined by the rendering frame rate (fps) and the motion duration $T$ (in seconds), yielding $N = \text{fps} \times T$ time steps.
Thus, each demonstration produces $N$ tuples of (hand pose, hand-camera image, third-person image).
In our experiments, we set $\text{fps}=30$ and the motion length to 10 seconds, resulting in $N=300$ time steps per demonstration.

For visual observations, both camera views are rendered by projecting point clouds onto images.
The hand camera (egocentric) view includes only the floor plane, the object, and (for place and push data) the floor marker described later, while the hand itself is not rendered.
The third-person camera is randomly placed above the floor plane and oriented toward the midpoint of the task-relevant entities (i.e., the object and the hand cube), ensuring that both are captured in the initial frame image.
Since we do not assume any specific robot morphology during this synthetic data generation process, we render the hand as a simple cube in third-person view.
This abstraction allows the data to remain robot-agnostic while still making the hand pose visually observable.
Accordingly, the third-person view contains the floor, the object, the marker (when present), and the hand represented as a cube.
All images are rendered as sparse binary projections on a black background, with visible point-cloud pixels drawn in white using a small, fixed footprint ($2\times2$ pixels) for clarity.
In the following, we describe the data generation method for each action.

\subsubsection{Pick}
Pick data represents the motion in which the hand moves to a position where it can grasp the object. 
The hand pose at time $t$ is denoted as $\mathbf{T}(t)=(\mathbf{p}(t), \mathbf{R}(t))$, where $\mathbf{p}(t)$ is the position and $\mathbf{R}(t)=[\mathbf{r}(t), \mathbf{u}(t), \mathbf{f}(t)]$ is the rotation matrix composed of right, up, and forward unit vectors.
The trajectory moves the hand from the initial pose $\mathbf{T}_0=(\mathbf{p}_0,\mathbf{R}_0)$ to the final grasp pose $\mathbf{T}_{\text{pick}}=(\mathbf{p}_{\text{pick}},\mathbf{R}_{\text{pick}})$ (see \figurename~\ref{fig:data-creation} pick).
The final grasp pose is defined using a simple geometric heuristic based on the object's bounding box.
We select a slender side of the object, choose the face closer to the current hand position, place the grasp point slightly offset outward from the center of that face, and orient the hand so that its forward direction points toward the center of the object.
The hand position during the motion is interpolated linearly.
\begin{equation}
\mathbf{p}(t) = (1-t)\,\mathbf{p}_0 + t\,\mathbf{p}_{\text{pick}}, \quad t\in[0,1].
\label{eq:linpos}
\end{equation}
For orientation, we interpolate the forward direction using spherical linear interpolation (SLERP)~\cite{SLERP}. 
Let $f_0$ be the initial forward direction and $f_{\text{pick}}$ the final forward direction. 
The interpolated forward direction at time $t$ is then defined as follows:
\begin{align}
\mathbf{f}(t) &= \frac{\sin((1-t)\theta)}{\sin \theta} \mathbf{f}_0 
     + \frac{\sin(t\theta)}{\sin \theta} \mathbf{f}_{\text{pick}},
\label{eq:slerp} \\
\theta &= \arccos(\mathbf{f}_0 \cdot \mathbf{f}_{\text{pick}}). 
\label{eq:theta}
\end{align}
The right and up directions are updated at each step to remain orthogonal to $\mathbf{f}(t)$, forming $\mathbf{R}(t)=[\mathbf{r}(t), \mathbf{u}(t), \mathbf{f}(t)]$.
Through this procedure, the hand poses in the pick action are generated as a time series. 
For the gripper value, we set 0 for all poses except the final one, where the value is set to 1.

\subsubsection{Place}
For the place data, the process begins by reproducing the pick motion used to grasp the object, followed by moving the object to a designated target location.
To visually indicate the object's target position and orientation, we draw a dense point cloud pattern on the plane and refer to it as a ``marker'' (\figurename~\ref{fig:data-creation}, bottom-right).
The marker is created by randomly rotating the object point clouds and projecting them from a top-down view.
Let this random rotation be denoted as $\mathbf{R}_{\text{rand}}$.
This rotation matrix $\mathbf{R}_{\text{rand}}$ determines how the object is projected onto the floor as a marker, thereby defining the object's final placement and orientation.
The task then consists of moving the grasped object to align with this marker.

The placement motion must satisfy two essential constraints:
1) The relative position between the hand and the object must be preserved during the motion, and 
2) The final pose of the object must perfectly overlap with the marker when viewed from above.
We define the hand pose at the end of the pick motion as
\begin{equation}
\mathbf{T}_{\text{pick}} = (\mathbf{p}_{\text{pick}}, \mathbf{R}_{\text{pick}}).
\label{eq:tstart}
\end{equation}
The target placement pose is given by
\begin{equation}
    \mathbf{T}_{\text{place}} = (\mathbf{p}_{\text{place}}, \mathbf{R}_{\text{place}}).
    \label{eq:tgoal}
\end{equation}
Here, $\mathbf{p}_{\text{place}}$ is computed using the center of the marker $\mathbf{c}_{\text{mark}}$.
The relative offset between the hand and the object centroid $\mathbf{c}_{\text{obj}}$, denoted by $\Delta \mathbf{p}$, is defined as
\begin{equation}
    \Delta \mathbf{p} = \mathbf{p}_{\text{pick}} - \mathbf{c}_{\text{obj}}.
    \label{eq:delta_p}
\end{equation}
The placement position is then expressed as

\begin{equation}
    \mathbf{p}_{\text{place}} = \tilde{\mathbf{c}}_{\text{mark}} + \mathbf{R}_{\text{rand}} \,\Delta \mathbf{p} .
\label{eq:p_place_lifted}
\end{equation}
$\tilde{\mathbf{c}}_{\text{mark}}$ is the position of the object's center when the object rests on the plane and its $x, y$ coordinates coincide with the center of the marker $\mathbf{c}_{\text{mark}}$.
This definition of the final position ensures that the hand maintains the same relative position to the object.
The target orientation is defined by
\begin{equation}
    \mathbf{R}_{\text{place}} = \mathbf{R}_{\text{rand}} \mathbf{R}_{\text{pick}}.
    \label{eq:r_place}
\end{equation}
As a result, after the placement motion, the same random rotation used for marker creation is applied to the object, ensuring that it perfectly overlaps with the marker in the top-down view.
To reproduce the lifting and lowering motions during placement, the hand’s position displacement is defined by a Bézier curve. 
Based on this hand position and $\Delta \mathbf{p}$, the moving object's trajectory is also determined. 
For orientation, SLERP ensures a smooth transition, similar to the picking motion.
The gripper value is 0 during the approach to the object, 1 during transport to maintain the grasp, and 0 again once the final placement pose is achieved.

3) Push: 
For the push data, we reproduce the motion of moving the object 
to a target position along the plane without lifting it. 
As in the place task, a marker is generated on the floor plane to define the target position. 
However, unlike the place task, the marker is created by parallel translation only, without additional rotation (\figurename~\ref{fig:data-creation} bottom right).

The push motion consists of an approach phase followed by a push phase. 
The pushing direction is defined as the horizontal component of the vector from the object center $\mathbf{c}_{\text{obj}}$ to the marker center $\mathbf{c}_{\text{mark}}$:
\begin{equation}
    \mathbf{v}_{\text{push}} = \text{Proj}_{xy}(\mathbf{c}_{\text{mark}} - \mathbf{c}_{\text{obj}}).
\end{equation}

In the approach phase, we first define the starting pose of the push. 
The pose just before contact is defined as
\begin{equation}
    \mathbf{T}_{\text{start}} = (\mathbf{p}_{\text{start}}, \mathbf{R}_{\text{start}}),
\end{equation}
with
\begin{equation}
    \mathbf{p}_{\text{start}} = \mathbf{c}_{\text{obj}} - d\,\mathbf{v}_{\text{push}}.
\end{equation}
The offset term $d\,\mathbf{v}_{\text{push}}$ prevents the hand cube from overlapping with the object point cloud. The scale factor \(d\) determines the hand–object offset along the pushing direction. The forward axis at this pose is aligned with the pushing direction,
\begin{equation}
    \mathbf{f}_{\text{start}} = \frac{\mathbf{v}_{\text{push}}}{||\mathbf{v}_{\text{push}}||},
\end{equation}
while the up axis is fixed to the world vertical $\mathbf{u}_{\text{world}}$, and the right axis is defined as $\mathbf{r}_{\text{start}} = \mathbf{u}_{\text{world}} \times \mathbf{f}_{\text{start}}$, forming $\mathbf{R}_{\text{start}} = [\mathbf{r}_{\text{start}}, \mathbf{u}_{\text{world}}, \mathbf{f}_{\text{start}}]$. 
The hand pose is then interpolated from the initial pose $\mathbf{T}_0$ to $\mathbf{T}_{\text{start}}$ using linear interpolation for position and SLERP for orientation.

After reaching $\mathbf{T}_{\text{start}}$, the push phase begins. 
The final pose is defined as
\begin{equation}
    \mathbf{T}_{\text{push}} = (\mathbf{p}_{\text{push}}, \mathbf{R}_{\text{push}}),
\end{equation}
with
\begin{equation}
    \mathbf{p}_{\text{push}} = \tilde{\mathbf{c}}_{\text{mark}} - d\,\mathbf{v}_{\text{push}},
\end{equation}
and $\mathbf{R}_{\text{push}} = \mathbf{R}_{\text{start}}$. The relative pose between the hand and the object is preserved during this phase. 
Thus, the hand and the object translate together along $\mathbf{v}_{\text{push}}$ while maintaining a constant orientation whose up axis coincides with the world vertical. 
For the push action, the gripper value remains 0 throughout.

\section{EXPERIMENTS}

In this section, we investigate how the constructed synthetic dataset influences learning convergence and success rates. Specifically, we conduct experiments in both simulation and real-world settings to evaluate the effectiveness of \rev{the proposed} pretraining.
To this end, we employ an imitation learning framework and compare the performance of models trained from scratch with that of models pre-trained on the synthetic dataset.
As the policy architecture, we adopt ACT~\cite{zhao2023learningfinegrainedbimanualmanipulation}.
ACT takes as input a hand camera image, a third-person camera image, and the current robot hand pose, and predicts the hand pose at the next time step.
Across all experiments, the hand pose is represented using a unified 8-dimensional vector consisting of the position coordinates, a quaternion, and the gripper opening value:
$
(x, y, z, \mathbf{q}, \text{gripper}),
$
where $\mathbf{q} \in \mathbb{R}^4$ denotes the quaternion.
This consistent representation ensures compatibility between synthetic pretraining data and task-specific data learning in both simulated and real-world environments.
\rev{To reduce the possibility that tasks are solved without using visual information}, we randomize the object positions in every evaluation trial across both simulated and real-world environments.
The details of experimental setups and results for each setting are described in the following subsections.

\begin{figure}[!t]
    \centering
    \includegraphics[width=0.99\columnwidth]{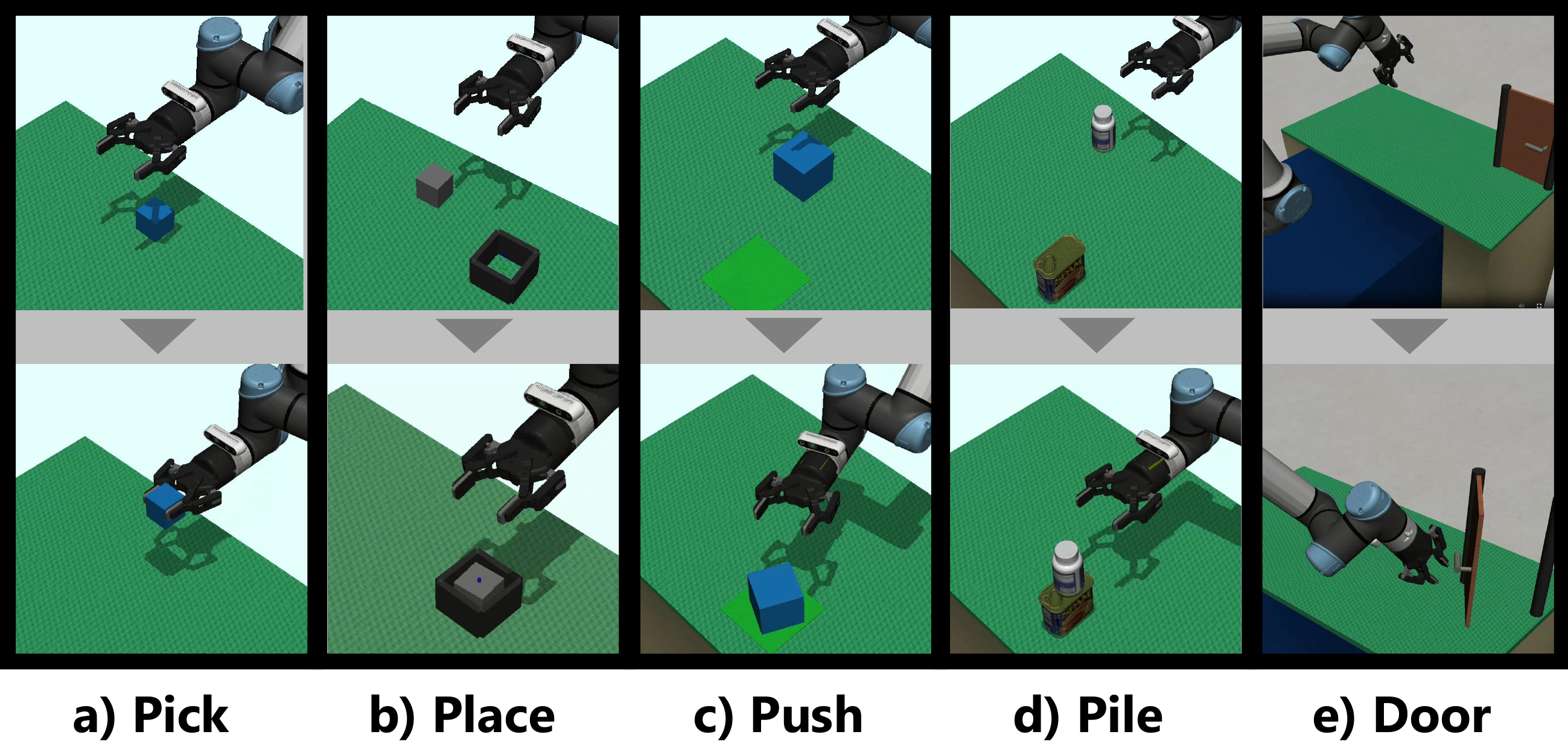}
    \caption{\textbf{Simulation Task Overview}. We conducted experiments on five manipulation tasks: a) pick, lifting a cube, b) place, putting a cube inside an enclosure, c) push, moving a cube to a colored area without lifting it, d) pile, placing a bottle on a can, and e) door, opening a door.}
    \label{fig:tasks}
\end{figure}
\begin{figure*}[t]
    \centering
    \includegraphics[width=0.95\textwidth]{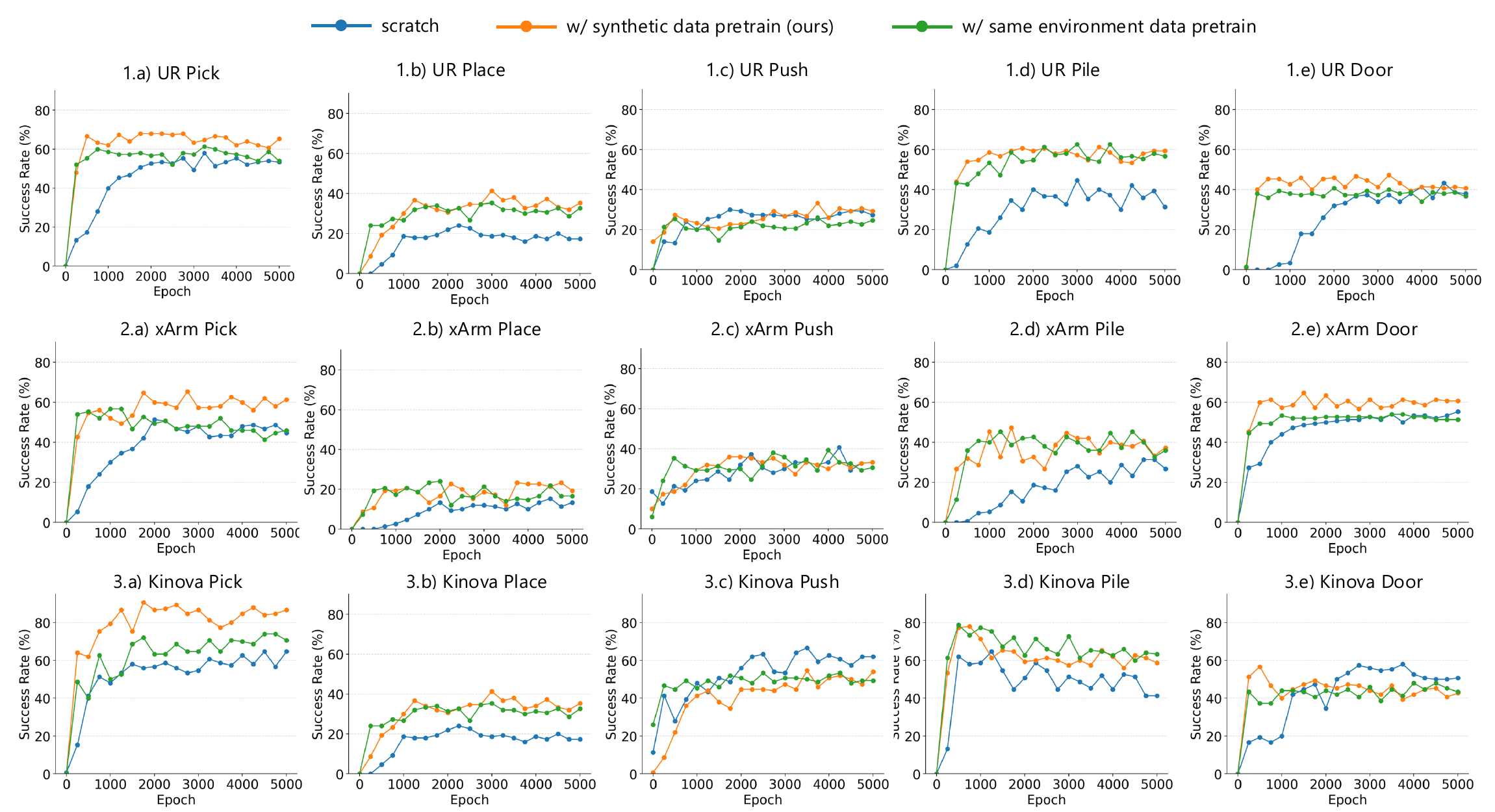}
    \caption{\textbf{Comparison of success rates for each robotic task in simulation experiment.} We compare three models: one trained from scratch, one pretrained on synthetic data, and one pretrained on data similar to the downstream task. Every 250 epochs, each model is evaluated over 50 test runs across three different random seeds, and the resulting mean success rates are plotted.}
    \label{fig:success_rate}
\end{figure*}

\subsection{Simulation Experiments}
\subsubsection{Experimental Setup}
\label{sec: setting}

\rev{For simulation experiments, we use RoboManipBaselines~\cite{murooka2025robomanipbaselinesunifiedframeworkimitation} as the software framework for data collection, training, and policy deployment.} To evaluate whether the proposed dataset provides pretraining benefits independent of robot morphology, we employed three types of robots: UR5e, xArm7, and KinovaGen3.
For each robot, five downstream manipulation tasks were prepared. 
An overview of these tasks is shown in \figurename~\ref{fig:tasks}.
For each task, 20 episodes were collected as 
training data.
We evaluated the effect of the proposed pretraining method across three training conditions:

\noindent\textbf{Scratch}~
The model was trained only on the 20 task-specific datasets without any pretraining.

\noindent\textbf{Synthetic Data Pretraining}~
The model was pretrained using the proposed synthetic dataset. 
This dataset consisted of 200 episodes for each of three primitive actions (pick, place, and push), for a total of 600 episodes. 
The synthetic data were converted to the same 8-dimensional hand-pose representation used for robot execution. We conducted pretraining on this synthetic dataset for 2000 epochs and then fine-tuned it on \rev{the 20 task-specific demonstrations}.

\noindent\textbf{Same Environment Data Pretraining}~
For comparison, we also performed pretraining using a dataset collected in the same MuJoCo environment as the downstream tasks. 
This dataset was collected using the UR robot and included nine types of tasks, ranging from simple pick-and-place to deformable object manipulation and open/close motions. 
To ensure a fair comparison with the synthetic dataset, we subsampled the data to match the total number of episodes to 600. 
The environmental settings, such as background and table color, were identical to those used in the downstream tasks.
We also performed pretraining on this dataset for 2000 epochs, as with the synthetic data, and then fine-tuned it on \rev{the 20 task-specific demonstrations.}

\begin{table}[t]
\caption{Comparison of peak success rates among models trained from scratch, models pretrained with synthetic data, and models pretrained with same-environment data.}
\label{tab:simple-results}
\centering

\small
\setlength{\tabcolsep}{3pt}
\renewcommand{\arraystretch}{1.15}

\resizebox{\linewidth}{!}{%
\begin{tabular}{llcccccc}
\toprule
Robot & Training Method & Pick & Place & Push & Pile & Door & Avg. \\
\midrule

\multirow{3}{*}{UR}
& Scratch
& $58.0 \pm 2.0$ & $24.0 \pm 7.0$ & $30.0 \pm 4.0$ & $44.7 \pm 5.7$ & $43.3 \pm 2.7$ & $40.0 \pm 2.4$ \\
& w/ Synthetic data (ours)
& \textbf{68.0 $\pm$ 9.2} & \textbf{41.3 $\pm$ 4.4} & \textbf{33.3 $\pm$ 3.7} & $61.3 \pm 1.8$ & \textbf{47.3 $\pm$ 3.3} & \textbf{50.3 $\pm$ 0.9} \\
& w/ Same environment data
& $61.3 \pm 4.4$ & $35.3 \pm 0.7$ & $26.0 \pm 3.1$ & \textbf{62.7 $\pm$ 4.7} & $40.7 \pm 1.8$ & $45.2 \pm 1.0$ \\
\midrule

\multirow{3}{*}{xArm}
& Scratch
& $51.3 \pm 6.7$ & $15.3 \pm 5.2$ & \textbf{40.7 $\pm$ 10.0} & $31.3 \pm 6.4$ & $55.3 \pm 1.3$ & $38.8 \pm 3.3$ \\
& w/ Synthetic data (ours)
& \textbf{65.3 $\pm$ 2.7} & $23.3 \pm 4.8$ & $36.0 \pm 4.2$ & \textbf{47.3 $\pm$ 4.7} & \textbf{64.7 $\pm$ 1.3} & \textbf{47.3 $\pm$ 1.0} \\
& w/ Same environment data
& $56.7 \pm 4.1$ & \textbf{24.0 $\pm$ 3.1} & $39.3 \pm 0.7$ & $45.3 \pm 8.4$ & $54.0 \pm 3.1$ & $43.9 \pm 2.4$ \\
\midrule

\multirow{3}{*}{Kinova}
& Scratch
& $64.7 \pm 11.4$ & $8.7 \pm 1.3$ & \textbf{66.7 $\pm$ 4.7} & $64.7 \pm 13.5$ & \textbf{58.0 $\pm$ 1.2} & $52.5 \pm 3.7$ \\
& w/ Synthetic data (ours)
& \textbf{90.7 $\pm$ 3.5} & \textbf{24.0 $\pm$ 1.2} & $54.7 \pm 5.2$ & $78.0 \pm 8.0$ & $56.7 \pm 12.2$ & \textbf{60.8 $\pm$ 0.2} \\
& w/ Same environment data
& $74.0 \pm 2.3$ & $16.7 \pm 7.0$ & $53.3 \pm 3.5$ & \textbf{78.7 $\pm$ 2.4} & $48.0 \pm 3.5$ & $54.1 \pm 1.5$ \\
\bottomrule
\end{tabular}%
}
\end{table}







\begin{table}[t]
  \caption{Effect of different data generation methods on the success rate. Values are the peak success rates averaged across three robots for each of the five tasks. The conditions are defined as follows: original synthetic data, black image data (all dataset images replaced with black), random motion data (random hand movement during data generation), no hand cube (hand cube rendering removed), and sphere only (all objects unified to the sphere).}
  \label{tab:ablation}
  \centering
  \setlength{\tabcolsep}{6pt}
  \renewcommand{\arraystretch}{1.2}

  \resizebox{\columnwidth}{!}{
  \begin{tabular}{lcccccc}
    \toprule
    Training Method & Pick & Place & Push & Pile & Door & Avg. \\
    \midrule
    Original synthetic data 
      & \textbf{74.7 $\pm$ 2.3} &  \textbf{29.6$\pm$3.3} & $41.3 \pm 3.1$ 
      & 62.2$ \pm $4.2 & 56.2 $\pm$ 4.2 & \textbf{52.8 $\pm$ 0.5} \\

    Black image data
      & $57.1 \pm 3.8$ & $22.9 \pm 1.5$ & $42.7 \pm 2.0$ 
      & $58.0 \pm 1.8$ & $53.8 \pm 1.6$ & $46.9 \pm 1.4$ \\

    Random motion data
      & $63.8 \pm 3.4$ & $24.9 \pm 1.6$ & \textbf{47.1 $\pm$ 2.1}
      & $61.3 \pm 1.4$ & $52.4 \pm 1.6$ & $49.9 \pm 1.2$ \\

    No hand cube data
      & $65.1 \pm 2.8$ & 28.4 $\pm$ 1.4 & $45.6 \pm 2.3$
      & 62.2 $\pm$ 3.2 & \textbf{56.4 $\pm $3.5} & $51.6 \pm 0.7$ \\

    Sphere only data
      & $70.7 \pm 3.2$ & 24.7 $\pm$ 2.3 & 42.9 $\pm$ 3.1
      & \textbf{68.2 $\pm$ 1.6} & $52.2 \pm 1.8$ & $51.7 \pm 0.6$ \\

    \bottomrule
  \end{tabular}}
\end{table}

\subsubsection{Results and Discussion}
To evaluate the effectiveness of the proposed pretraining method, we conducted experiments addressing the following two key questions. \\
\noindent\textbf{Q1}: How does the proposed synthetic dataset affect the convergence speed of task learning and task success rate?\\
\rev{\noindent\textbf{Q2}: Which design elements of the generated data contribute to the pretraining effect?}\\
In the following sections, we present and analyze the experimental results in detail, focusing on these two key questions.

\paragraph{\textbf{Q1}: How does the proposed synthetic dataset affect the convergence speed of task learning and task success rate}

First, to evaluate the effect of synthetic data on convergence speed, we conducted experiments using three types of robots and five tasks. 
Figure~\ref{fig:success_rate} shows the success rate under three conditions: training from scratch, pretraining with our synthetic data, and pretraining with \textit{same environment data} (see \textit{Experimental Setup}). 
The success rate was measured by saving the model every 250 epochs up to 5,000 epochs across three different seeds. 
For each seed, 50 trials were conducted, and the average success rate across these seeds is presented in Figure~\ref{fig:success_rate}.

\consistencyrev{In Fig.~\ref{fig:success_rate}, synthetic-data pretraining shows higher success rates at several early fine-tuning checkpoints than training from scratch, although the pattern varies by task.}
\consistencyrev{A similar early-stage trend is visible for same-environment data pretraining on some tasks.}
\consistencyrev{Table~\ref{tab:simple-results} reports peak success rates selected over the evaluated checkpoints, not success rates at a common fixed epoch. These peak values therefore do not by themselves quantify convergence speed.}
\consistencyrev{As shown in Table~\ref{tab:simple-results}, the proposed method has the highest five-task mean of peak success rates for each robot, compared with both Scratch and Same Environment Data Pretraining.}
\consistencyrev{The latter yields lower five-task mean peak success rates than the proposed method for all three robots despite using data collected in a similar environment.}
\consistencyrev{For Kinova, Same Environment Data Pretraining improves the five-task mean peak success rate over Scratch by only 1.6 percentage points.}
One possible reason is that the \textit{same environment data} was collected using a different robot (UR).
\consistencyrev{In contrast, Synthetic Data Pretraining improves the five-task mean peak success rate over Scratch by 10.3 percentage points for UR, 8.5 percentage points for xArm, and 8.3 percentage points for Kinova.}
\consistencyrev{These values describe improvements in peak performance, not reductions in training time. Both pretrained conditions also involve 2,000 pretraining epochs in addition to task-specific fine-tuning; a total-cost comparison requires accounting for this stage.}
\rev{Per-task results are nevertheless mixed. The proposed pretraining does not improve over training from scratch on Push for xArm and Kinova, nor on Door for Kinova. The reported gains should therefore be read as an improvement of the average over the task set, rather than as a uniform improvement on every task.}
\rev{Consistent with this,} the push task shows limited improvement under both pretraining settings.
\rev{One possible explanation is that} the difficulty of push tasks lies in maintaining a stable hand-object relationship amid friction and slippage, rather than merely in acquiring geometric representations.
While pretraining provides better spatial priors, the policy must still adapt to the unpredictable dynamics of contact-rich motion during fine-tuning.

\paragraph{\rev{\textbf{Q2}: Which design elements of the generated data
contribute to the pretraining effect}}
To validate the effectiveness of our synthetic data design, we conducted an ablation study by modifying key components of the data-generation process and comparing the resulting pretraining effects.
For this evaluation, we used the average values across three robots for each of the five tasks.
In total, we prepared five types of synthetic datasets: the proposed original synthetic data and four variants in which specific design elements were altered. The four modified datasets are defined as follows:

\noindent\textbf{Black image data}~ The images from both the hand camera and the third-person camera were replaced with all-black images, leaving only the time-series hand-pose information available.
    
\noindent\textbf{Random motion data}~ Instead of generating structured primitive actions (pick, place, and push), the hand was moved randomly. The corresponding hand poses, hand camera, and third-person camera images were recorded.
    
\noindent\textbf{No hand cube data}~ The cube representing the hand pose was not rendered in the third-person camera view, making the hand's visual state unavailable from that perspective.
In contrast, the hand camera images remained unchanged.

\noindent\textbf{Sphere-only data}~ Instead of using a diverse range of objects, we unified the objects in the synthetic data to spheres only and recorded the observations and hand poses during primitive actions.

For each dataset, we generated 600 episodes in the same manner as the original synthetic data. 
All models were pretrained for 2000 epochs. 
We then evaluated the task success rates. Table~\ref{tab:ablation} represents the peak success rate achieved for each task.
\consistencyrev{For reference, the Scratch results in Table~\ref{tab:simple-results}, averaged equally over the three robots and five tasks, give a mean peak success rate of 43.8\%, compared with 52.8\% for the original synthetic data in Table~\ref{tab:ablation}. On Push, however, the corresponding values are 45.8\% and 41.3\%, respectively.}

\rev{In terms of the average success rate across the five tasks}, performance increases in the following order: black image data, random motion data, no-hand cube data, sphere-only data, and original synthetic data.
The black image data removes all visual information, while the random motion data eliminates meaningful motion structure in the recorded trajectories.
\consistencyrev{The lower five-task averages for these two variants are consistent with contributions from visual observations and structured motion. However, the current ablations do not isolate a geometry-specific prior from generic benefits of additional pretraining.}
\rev{This ordering describes the average and is not monotone for every individual task. On Push in particular, the original data is the weakest of the five conditions and random motion data the strongest, which is consistent with the observation above that Push depends on contact dynamics that the proposed geometric pretraining does not model.}

Furthermore, the original synthetic data achieves higher average success rates than the no-hand cube data.
This result suggests that although the hand camera images provide visual cues about the hand–object relationship, the additional explicit visualization in the third-person images \rev{provides a small additional gain (52.8 vs. 51.6 on average)}.

Sphere-only data achieved success rates closest to the original data\rev{, and the difference in the overall average is small (52.8 vs. 51.7)}.
However, for tasks like Door, where the manipulated object differs significantly from a sphere, the original data outperformed it \rev{by a larger margin (56.2 vs. 52.2)}.
\rev{These results suggest that object shape diversity has a limited effect on average, but a relatively larger effect on tasks whose objects deviate from the shapes seen during pretraining.}

\subsection{Real Robot Experiments}
\subsubsection{Experimental Setup}
\begin{figure}[!t]
    \centering
    \includegraphics[width=0.99\columnwidth]{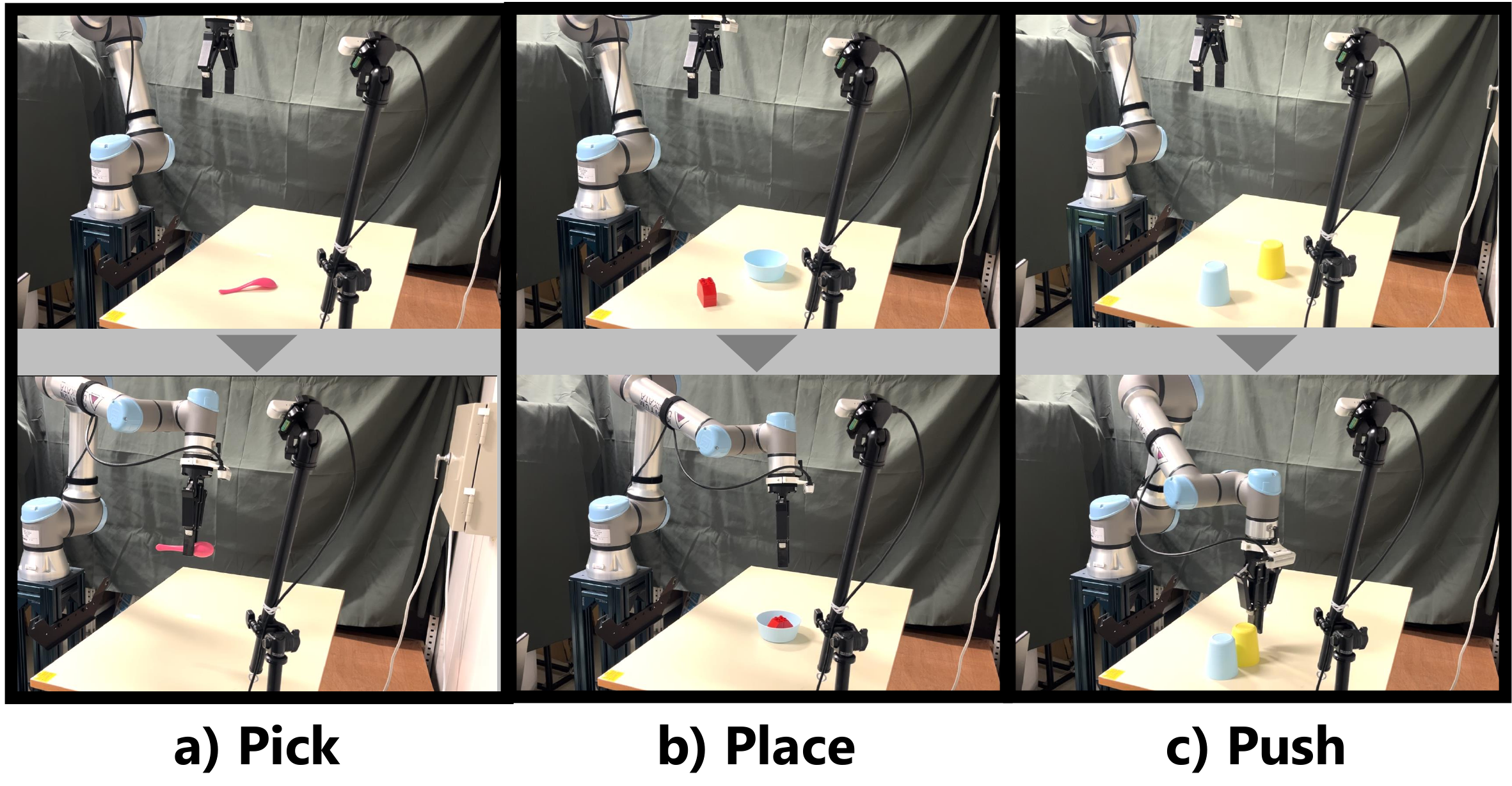}
    \caption{\textbf{Real Task Overview}. We conducted experiments on three manipulation tasks: a) pick, lifting a spoon, b) place, putting a block inside a plate, c) push, moving a cup to another cup.}
    \label{fig:real_tasks}
\end{figure}

\begin{figure}[t]
    \centering
    \includegraphics[width=0.99\columnwidth]{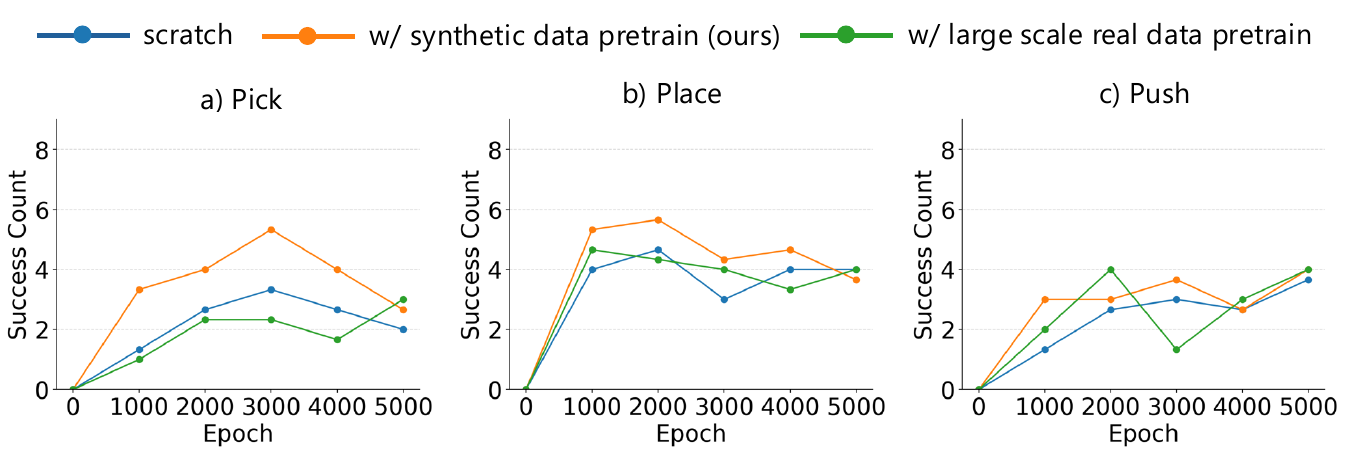}
    \caption{\textbf{Comparison of success counts for each robotic task in a real robot experiment.} We compare three training conditions: Scratch, Synthetic Data Pretraining (ours), and Large-Scale Real-Data Pretraining. Every 1,000 epochs, each model is evaluated over 10 test runs across three different random seeds, and the mean success count is plotted.}
    \label{fig:real_robot}
\end{figure}

To evaluate whether the proposed synthetic dataset provides pretraining benefits in real-world settings, we conducted experiments using a UR robot. 
Figure~\ref{fig:real_tasks} illustrates an overview of the tasks. We defined three tasks: pick, place, and push.
\consistencyrev{For each task, Fig.~\ref{fig:real_robot} shows three training conditions. We evaluated the following three training conditions:}

\noindent\textbf{Scratch}~
The model was trained from scratch using only task-specific data. 
For each task, 30 real-world episodes were collected and used for training.

\noindent\textbf{Synthetic Pretrain}~
The model was first pretrained using the proposed synthetic dataset.
As in the simulation experiments, 2,000 epochs of pretraining were conducted using 600 synthetic data episodes, followed by fine-tuning with task-specific data\rev{.}

\noindent\textbf{Large-scale Real Pretrain}~
The model was pretrained using large-scale real-world data. Specifically, we used approximately 60k successful episodes from the DROID dataset~\cite{khazatsky2024droid}.
DROID is a large-scale real-world teleoperation dataset collected with a Franka Panda robot, whose main strength is its substantial environmental diversity while maintaining a consistent and standardized data format.
For training, we used hand and third-person camera images and hand pose information, following the same setup as in the synthetic data, and performed two epochs of pretraining.


\subsubsection{Results and Discussion}
Figure \ref{fig:real_robot} shows the average success counts across three different random seeds. Models were evaluated every 1,000 epochs up to 5,000 epochs, with 10 trials per saved model. For the pick-and-place tasks, the model pretrained with the proposed synthetic data achieved \rev{a higher peak success count than training from scratch}. In addition, the model pretrained with the proposed method outperformed training from scratch at most of the evaluated learning stages, measured at 1,000-epoch intervals.

\section{CONCLUSIONS}
We introduced a \rev{framework that automatically generates purely geometric} data in a simple spatial environment, which we then use for pretraining followed by fine-tuning.
\rev{The proposed approach aims to provide useful geometric priors, in particular, a representation of the relative pose between the hand and the object, so that a policy can be trained efficiently from a small number of task demonstrations.}
\consistencyrev{In simulation, the proposed pretraining improves the five-task mean of peak success rates for all three robots, although the benefit is not observed on every individual task. The real-robot curves suggest an early-stage advantage on some tasks. }
\rev{\consistencyrev{The ablation results are consistent with contributions from visual observations and structured trajectories, but do not establish that these factors account for most of the gain over Scratch or rule out generic warm-start effects. The hand-cube and object-shape variants show smaller changes in the five-task average.} We do not claim robustness to arbitrary cluttered environments. The purpose of the proposed pretraining is not to model realistic environments but to provide a geometric initialization that can later be adapted during fine-tuning. We also note that all experiments in this work use ACT as the policy architecture, and whether the same benefit transfers to other visuomotor policies remains an open question.} As future work, the proposed \rev{data} generation framework could be extended to support dual-arm robotic systems and tasks involving interactions among multiple objects.

\section*{Acknowledgment}
This work was supported by the National Institute of Advanced Industrial Science and Technology (AIST) Policy Budget Project, ``Research and Development of Foundation Models for Generative AI in the Physical Domain.''

\bibliographystyle{IEEEtran}
\bibliography{ref}

\end{document}